\documentclass[runningheads]{llncs}

\usepackage[utf8]{inputenc}
\usepackage{graphicx}
\usepackage{float}
\usepackage{footnote}
\usepackage{url}
\usepackage{listings}
\usepackage{booktabs}
\usepackage{array}
\usepackage{lscape}
\usepackage{multirow, makecell}
\usepackage{footnote}

\begin{document}

\title{\textit{TNNT}: The Named Entity Recognition Toolkit}
\titlerunning{\textit{TNNT}: The Named Entity Recognition Toolkit}

\author{
    Sandaru Seneviratne\inst{1,4}\orcidID{0000-0001-8320-5084} \and
    Sergio J. Rodríguez Méndez\inst{1,2}\orcidID{0000-0001-7203-8399} \and
    Xuecheng Zhang\inst{1,4} \and
    Pouya G. Omran\inst{1,3}\orcidID{0000-0002-4473-3877} \and
    Kerry Taylor\inst{1,4}\orcidID{0000-0003-2447-1088} \and
    Armin Haller\inst{1,4}\orcidID{0000-0003-3425-0780}
}

\authorrunning{Seneviratne et al.}

\institute{
    Australian National University, Canberra ACT 2601, AU\and
    \email{Sergio.RodriguezMendez@anu.edu.au} \and
    \email{P.G.Omran@anu.edu.au} \and
    \email{\{firstname.lastname\}@anu.edu.au}\\
    \url{https://cs.cecs.anu.edu.au/} 
}

\date{\today}
\maketitle
\vspace{-1mm}

\begin{abstract}
Extraction of categorised named entities from text is a complex task given the availability of a variety of Named Entity Recognition (NER) models and the unstructured information encoded in different source document formats. Processing the documents to extract text, identifying suitable NER models for a task, and obtaining statistical information is important in data analysis to make informed decisions.  This paper presents\footnote{The manuscript follows guidelines to showcase a demonstration that introduces an overview of how the toolkit works: input document set, initial settings, processing, and output set. The input document set is artificial in order to show various toolkit capabilities.} \textit{TNNT}, a toolkit that automates the extraction of categorised named entities from unstructured information encoded in source documents, using diverse state-of-the-art Natural Language Processing (NLP) tools and NER models.  \textit{TNNT} integrates 21 different NER models as part of a Knowledge Graph Construction Pipeline (KGCP) that takes a document set as input and processes it based on the defined settings, applying the selected blocks of NER models to output the results.  The toolkit generates all results with an integrated summary of the extracted entities, enabling enhanced data analysis to support the KGCP, and also, to aid further NLP tasks.

\end{abstract}


\section{Introduction}
NER is a major component in NLP systems to extract information from unstructured text. Recent advances in Deep Learning and NLP have resulted in the availability of a large number of NER tools and models for use which have enabled NER of different categories from text.  However, given the existence of a wide range of document formats, extracting information is difficult considering the pre-processing required prior to using NER tools and the challenge of identifying which models to use. Having a system which provides easy processing of different document formats, easy selection of different models or tools, an integrated summary of the entities identified by the models and an API which enables basic functionalities to access the results of the models can enhance data analysis, accurate decisions and provide a thorough overview of the data used.

This paper introduces \textit{TNNT}\footnote{The project's URI is \url{https://w3id.org/kgcp/MEL-TNNT}.  All resources along with demo videos are available at this address.}.  Its main goal is to automate the extraction of categorised named entities from the unstructured information encoded in the source documents, using recent state-of-the-art NLP-NER tools and models.  \textit{TNNT} is integrated with the ``Metadata Extractor \& Loader" (\textit{MEL}) which implements a set of methods to extract metadata (and content-based information) from various file formats \cite{MEL}.

\section{Core Features}

\begin{table}[!t]
\centering
\caption{Tools and models integrated in \textit{TNNT}}\label{models} 
\tiny
\begin{tabular}{|p{0.5cm}|p{3cm}|p{8cm}|} 
\hline
\thead{\#} & \thead{Tool} & \thead{Number of Models}\\ 
\hline
1 &NLTK \cite{loper2002nltk} & 1 \\
2 &spaCy\footnote{\url{https://spacy.io/}} & 3 (en\_core\_web\_sm, en\_core\_web\_md, en\_core\_web\_lg)\\
3& Stanford NER \cite{manning2014stanford}& 3 (3-class model, 4-class model, 7-class model) \\
4 &Stanza \cite{qi-etal-2020-stanza} & 1 \\
5 &Flair \cite{akbik2019flair} & 5 (ner, ner-fast, ner-pooled, ner-ontonotes, ner-ontonotes-fast) \\
6 &Allen NLP \cite{gardner2018allennlp} & 2 (Elmo-based NER, Fine-grained NER)\\
7 &Polyglot \cite{polyglot} & 1 \\
8& Deeppavlov \cite{burtsev2018deeppavlov} & 4 (ner\_conll2003, ner\_ontonotes, ner\_conll2003\_bert, ner\_ontonotes\_bert)\\
9 &NER based on BERT \cite{devlin2018bert} & 1\\
\hline
\end{tabular}
\end{table}

\textit{TNNT} integrates 21 different NER models from 9 state-of-the-art NLP tools (Table~\ref{models}). These 21 models can identify up to 18 categories (Table~\ref{categories}) of named entities in text. The system is capable of processing different models sequentially based on the input settings (processing blocks) defined by the user.  All textual content extracted by \textit{MEL} is processable for \textit{TNNT} with a hybrid processing data flow, either from/to a document store\footnote{Currently, \textit{TNNT} only supports CouchDB (\url{https://couchdb.apache.org/})} or via direct processing from files.

For data analysis tasks, TNNT keeps general statistics of the models and generates an integrated summary of all the identified entities. The results are JSON\footnote{\url{https://www.json.org/}} files (one for each processed source document) with the list of models, categories, and identified entities.  For each recognised entity, the toolkit retrieves its context information and the start/end index in the document text\footnote{Sample results can be found at the project's \texttt{w3id} URI.}.  Table~\ref{results_table} gives an overview of the results obtained using some of the models for two publicly available datasets: CONLL 2003\footnote{\url{https://www.clips.uantwerpen.be/conll2003/ner/}} and NIST IE-ER\footnote{\url{https://github.com/juand-r/entity-recognition-datasets}}. 


\begin{table}[!t]
\centering
\caption{\textit{TNNT} results from some NER models for two public datasets}\label{results_table} 
\tiny
\begin{tabular}{|>{\centering\arraybackslash}p{1.6cm}|>{\centering\arraybackslash}p{1.5cm}|>{\centering\arraybackslash}p{1.5cm}|p{7.5cm}|} 
\hline
\thead{Model} & \thead{Dataset} & \thead{Exec. Time\\(seconds)} & \thead{Number of Recognised Entities}\\
\hline
Stanford-3 class model & CONLL 2003 & 17.16 & location:2165, organisation:2586, person:2726 (Total = 7477) \\
& NIST IE-ER & 7.55 & location:403, organisation:431, person:831 (Total = 1665)\\
\hline
Spacy-encore\_web\_md& CONLL 2003& 36.82 & location:112, organisation:2047, person:2921, NORP:931, FAC:90, GPE:3015, product:62, event:221, work\_of\_art:43, law:11, language:21, date:2890, time:266, percent:138, money:129, quantity:141, ordinal:367, cardinal:3469 (Total = 16874)  \\
& NIST IE-ER  & 14.55 & location:102, organisation:1184, person:1675, NORP:380, FAC:57, GPE:707, product:41, event:37, 
work\_of\_art:53, law:10, language:7, date:771, time:112, percent:48, money:23, quantity:37, ordinal:118, cardinal:609 (Total = 5971) \\
\hline
BERT-based& CONLL 2003  & 1245.66 & location:2312, organisation:2450, person:2723, miscellaneous:1381 (Total = 8866)\\
& NIST IE-ER & 662.27 & location:792, organisation:806, person:1269, miscellaneous:672 (Total = 3539)\\
\hline
\end{tabular}
\end{table}

Additionally, a built-in RESTful API provides basic functions to browse the results and to complement them by performing other NLP tasks, such as part-of-speech tagging, dependency parsing, and co-reference resolution.  These functionalities along with the comprehensive information provided by \textit{TNNT}, facilitate the understanding of the models and data used for NLP and KGCP tasks.

\begin{table}[!t]
\centering
\caption{Categories identified by the models integrated in \textit{TNNT}}\label{categories} 
\tiny
\begin{tabular}{|p{2cm}|p{8cm}|}
\hline
\thead{Category} & \thead{Description}\\
\hline
PERSON  &People, including fictional \\
NORP  &Nationalities or religious or political groups \\
FAC &Buildings, airports, highways, bridges, etc.\\
ORG &Companies, agencies, institutions, etc.\\
GPE &Countries, cities, states\\
LOCATION &Non-GPE locations, mountain ranges, bodies of water\\
PRODUCT &Objects, vehicles, foods, etc. (Not services.) \\
EVENT &Named hurricanes, battles, wars, sports events, etc.\\
WORK\_OF\_ART &Titles of books, songs, etc.\\
LAW &Named documents made into laws\\
LANGUAGE &Any named language\\
DATE &Absolute or relative dates or periods\\
TIME &Times smaller than a day.\\
PERCENT &Percentage, including “\%“\\
MONEY &Monetary values, including unit\\
QUANTITY &Measurements, as of weight or distance\\
ORDINAL &“first”, “second”, etc\\
CARDINAL &Numerals that do not fall under another type\\
\hline
\end{tabular}
\vspace{-4mm}
\end{table}

\section{Architecture}

\begin{figure}
\centering
\includegraphics[width=0.7\textwidth]{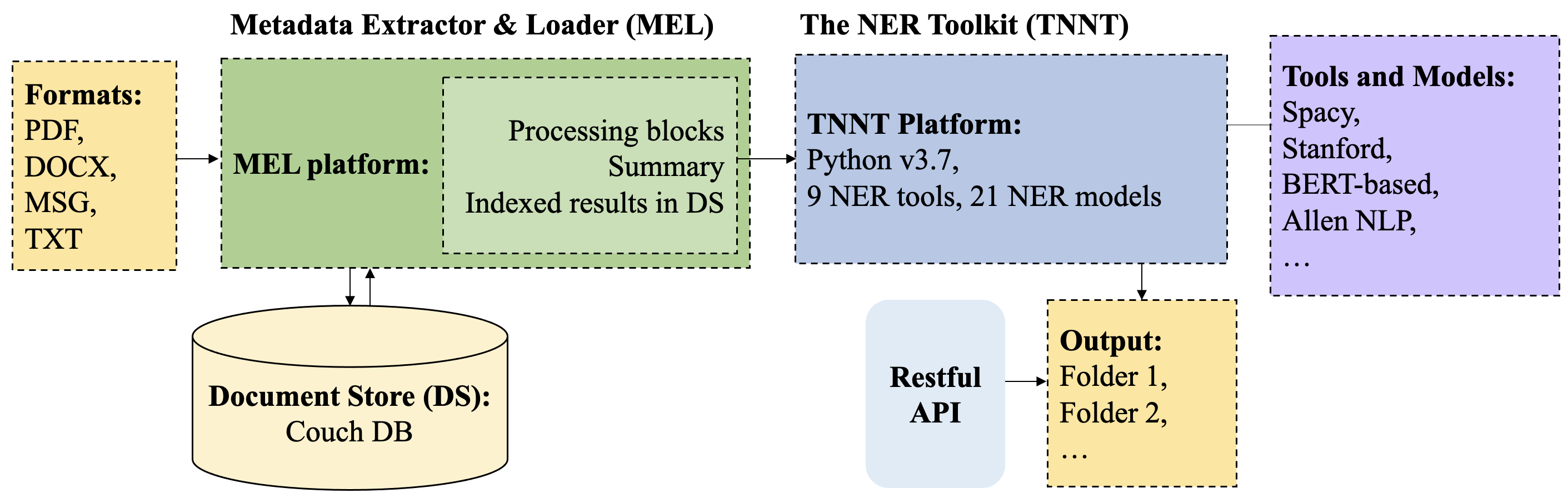}
\caption{\textit{TNNT} Architecture}\label{overview}
\vspace{-5mm}
\end{figure}
\textit{TNNT} has been fully integrated with \textit{MEL} (Figure \ref{overview}).  \textit{MEL} settings establish the way \textit{TNNT} will process some specific blocks sequence of NER models for the input dataset (either from content stored on a document store or from a direct document processing immediately after metadata extraction). More design details can be found at the project's \texttt{w3id} URI.


\section{Conclusions and Future Work}
\textit{TNNT} provides a simple mechanism to extract categorised named entities from unstructured data using a diverse range of state-of-the-art NLP tools and NER models.  This tool is still in its early stages of development.  It has been tested using different document formats and datasets as part of the ``Australian Government Records Interoperability Framework" (AGRIF) project.  There are ongoing plans to integrate more NER tools and models into the architecture along with continuing evolve the RESTful API with complementary NLP tasks to enrich the NER results, in order to support KGCP tasks.  The major contributions of this tool are: (1) the ability to process different source document formats for NER; (2) the availability of 21 different state-of-the-art NER models integrated in one system, enabling easy selection of models for NER; (3) the provision of an integrated summary of the results from different models; and (4) a RESTful API that enables easy access to the NER results from the models.

\footnotesize
\bibliographystyle{splncs04}
\bibliography{ref}

\end{document}